\documentclass[10pt,twocolumn,letterpaper]{article}

\usepackage{cvpr}              

\usepackage{graphicx}
\usepackage{amsmath}
\usepackage{amssymb}
\usepackage{booktabs}
\usepackage[accsupp]{axessibility}

\usepackage[pagebackref,breaklinks,colorlinks]{hyperref}

\usepackage[capitalize]{cleveref}
\crefname{section}{Sec.}{Secs.}
\Crefname{section}{Section}{Sections}
\Crefname{table}{Table}{Tables}
\crefname{table}{Tab.}{Tabs.}

\def\cvprPaperID{10165} 
\def\confName{CVPR}
\def\confYear{2023}

\newcommand{\bing}[1]{{\color{red}{\small\bf\sf [bing: #1]}}}

\begin{document}

\title{Dealing with Cross-Task Class Discrimination in Online Continual Learning}

\author{Yiduo Guo$^{1}$, Bing Liu$^{4}$, Dongyan Zhao$^{1,2,3}$\\ \fontsize{9.35}{12}\selectfont
$^{1}$Wangxuan Institute of Computer Technology, Peking University. 
$^{2}$BIGAI, Beijing, China. 
$^{3}$National Key Laboratory of \\\fontsize{9.35}{12}\selectfont General Artificial Intelligence.
$^{4}$Department of Computer Science, University of Illinois Chicago \\
{\tt\small yiduo@stu.pku.edu.cn, liub@uic.edu, zhaodongyan@pku.edu.cn}
}

\maketitle


\begin{abstract}

Existing continual learning (CL) research regards \textit{catastrophic forgetting} (CF) as almost the only challenge. This paper argues for another challenge in {class-incremental learning} (CIL), which we call \textbf{cross-task class discrimination} (CTCD),~i.e., how to establish decision boundaries between the classes of the new task and old tasks with no (or limited) access to the old task data.~CTCD is implicitly and partially dealt with by replay-based methods. A replay method saves a small amount of data (\textit{replay data}) from previous tasks. When a batch of current task data arrives, the system jointly trains the new data and some sampled replay data. The replay data enables the system to partially learn the decision boundaries between the new classes and the old classes as the amount of the saved data is small. However, this paper argues that the replay approach also has a \textbf{dynamic training bias} issue which reduces the effectiveness of the replay data in solving the CTCD problem.    
{\color{black}A novel optimization objective with a gradient-based adaptive method is proposed to dynamically deal with the problem in the online CL process.} Experimental results show that the new method achieves much better results in online CL. 


 \end{abstract}
 \section{Introduction}

\label{sec.intro}
Continual learning (CL) learns a sequence of tasks incrementally. 
This work focuses on the \textit{class incremental learning} (CIL) setting~\cite{masana2020class} in online CL. In CIL, each task consists of a set of unique classes, the sets of classes of any two different tasks are disjoint and the system has no access to the task information in testing. In online CL, the data comes gradually from a data stream. Whenever the small batch of data arrives, it is trained in one iteration. Thus, the data for each task is effectively trained in one epoch. 

Existing CL papers almost regard \textit{catastrophic forgetting} (CF) as the only issue for CL. In fact, CIL also has another major challenge.~When the system learns a new task, if no data from previous tasks is available, it has no way to establish decision boundaries between new classes and old classes in previous tasks. Even if there is no CF, the classification results will still be poor. We call this problem, \textbf{cross-task class discrimination} (CTCD). Those approaches that do not save any previous data, e.g., \textit{regularization-based} or \textit{orthogonal projection-based}, do not deal with CTCD. Replay-based methods implicitly deal with CTCD to some extent 
because such a method uses a memory buffer $\mathcal{M}$ to save a small amount of data (\textit{replay data}) from old tasks. When a small batch of current task data $X^{\textit{new}}$ arrives, the system jointly trains $X^{\textit{new}}$ and some sampled replay data $X^{\textit{buf}}$ from $\mathcal{M}$.  $X^{\textit{buf}}$ enables the system to partially learn the decision boundaries between the new classes and the old classes because the amount of the saved data is very small. 

Due to the limited replay data, the \textbf{training is biased}, which reduces its ability to solve the CTCD problem. To make matters worse, the training bias also changes as more tasks are learned. This paper first shows that the problem is reflected as \textbf{gradient imbalance} (GI) on logits, i.e., higher positive gradients than negative gradients \textit{on the logits} and vice versa. It further shows that GI is caused by two main issues. The first is \textbf{\textit{data imbalance}}.~Since the memory buffer size, the batch size of the new data $X^{\textit{new}}$, and the sampled data $X^{\textit{buf}}$ from the memory buffer are all fixed, if the system has learned many tasks, the average number of samples in each previous class in $X^{\textit{buf}}$ will be much smaller than that of each class in $X^{\textit{new}}$. This results in higher positive gradients than negative gradients on the logits of the previous classes leading to training bias and poor 
decision boundaries (or weak CTCD capability) between the classes of the new and old tasks. The second is \textbf{\textit{CL imbalance}}, i.e., CL training focuses more on the new samples (which are harder to train 
as they are new) than the replayed samples (which have been seen and trained many times before). This causes further GI. 
This imbalance is involved (see Sec.~\ref{sec.imbalance2} for details). 

{\color{black}Some existing works\cite{ahn2021ss,soutif2021importance} have tried to deal with data imbalance in offline CL. For example, SSIL\cite{ahn2021ss} separately calculates the cross-entropy loss of the new data and the replay data to mitigate data imbalance. 
But they are not from the gradient angle.} The second issue of GI is more complex and has not been attempted before. 

{\color{black}This paper proposes a novel method, called {GSA} (\textit{G}radient \textit{S}elf-\textit{A}daptation), to deal with GI (and CTCD) in online CL. GSA includes a new training objective and a   
{\textit{gradient-based self-adaptive loss}} to compensate for the GI.} The loss is dynamically controlled by two \textit{gradient rates} which automatically measure and adapt to the dynamic GI situation. {\color{black}The main contributions of this paper are:

(1) It deals with the CTCD problem in online CL and proposes a new optimization framework that decomposes the problem into cross-task classification and within-task classification (see Section~\ref{sec.proposed}).~In~\cite{kimtheoretical}, CTCD is called \textit{inter-task class separation}, but it uses an out-of-distribution based approach to dealing with the problem in offline CL. The paper uses a replay-based approach for online CL.} 

(2) It analyzes the CTCD problem from the gradient imbalance (GI) angle and finds two kinds of gradient imbalance (data imbalance and CL imbalance) (see Section~\ref{sec.analysis}). Based on the analysis, it proposes a \textit{gradient-based self-adaptive loss} to compensate for the GI.

(3) Experiments in both the disjoint and long-tail online CL settings show that GSA outperforms strong baselines by a large margin (see Section~\ref{sec.experiments}). 

\section{Related Work}
\label{sec.related}
There are many existing CL techniques. \textit{Regularization-based} approaches
penalize changes to important parameters of old tasks~\cite{kirkpatrick2017overcoming,
yu2020semantic,zhang2020class}.~\textit{Replay-based} approaches save some past data and replay them in new task training~\cite{chaudhry2019tiny,castro2018end,rebuffi2017icarl,Cyprien2019episodic,
zhao2021memory,korycki2021class,
yan2021framework,
wang2022memory}.~\textit{Generative replay} builds data generators to generate pseudo old data for replaying~\cite{Shin2017continual,
Seff2017continual,wu2018memory,Kemker2018fearnet,hayes2019remind,ostapenko2019learning}. \textit{Dynamic architectures} based approaches~\cite{
von2019continual,Serra2018overcoming,abati2020conditional,
Yoon2018lifelong,li2019learn,hung2019neurIPS,rajasegaran2019neurIPS,farajtabar2020orthogonal} overcomes CF by expanding or isolating parameters. Data augmentation has also been used to learn better features for CL recently~\cite{zhu2021class}. 


\textbf{Online CL and replay methods:} 
Existing online CL methods mainly use the replay approach. 
ER randomly samples the buffer data~\cite{chaudhry2019continual}. MIR chooses buffer data whose loss increases the most~\cite{aljundi2019online}. Shapley value theory is applied in ASER for  memory update/retrieval~\cite{shim2021online}. Knowledge distillation is employed in {DER++}~\cite{buzzega2020dark}. {NCCL} deals with CF by calibrating the network~\cite{yin2021mitigating}. Contrastive learning is used in {SCR}~\cite{mai2021supervised}. GDumb samples and stores class-balanced data in memory~\cite{prabhu2020gdumb}. However, it does not deal with CL imbalance identified in this paper. {GSS} diversifies the gradients of the samples in the buffer~\cite{aljundi2019gradient}. 
ER-AML~\cite{caccia2021reducing} avoids CF by shielding the learned representations from drastic adaptation to accommodate new classes.
OCM is based on mutual information~\cite{guo2022online}.  
Our work identifies two gradient imbalance issues and designed a new training strategy and a novel loss function to deal with them. 

\textbf{Data imbalance in CL}: {Several researchers have dealt with data imbalance in offline CL. 
SSIL~\cite{ahn2021ss} isolates the computation of softmax on previous and new classes to update the model.
LUCIR~\cite{hou2019learning} uses the cosine
normalization to calculate the predicted probability and a margin ranking loss to separate ground-truth old classes from new classes. \cite{castro2018end} adds an additional fine-tuning stage with a small learning rate and a balanced subset of samples. BiC~\cite{wu2019large} 
adds a bias correction layer. 
\cite{zhao2020maintaining} uses weight alignment
to correct the biased weights. 
CCIL~\cite{mittal2021essentials} applies a loss to balance intra-task and inter-task learning.
However, many of these algorithms \cite{zhao2020maintaining,castro2018end} need the full data of the current task to be available upfront and/or multiple training epochs to address the data imbalance issue, which are not suitable for online CL as online CL does not have the full training data of a task available when the task arrives in the data stream.} 

\textbf{Gradient-based CL methods}:
{GEM~\cite{Lopez2017gradient} and A-GEM \cite{chaudhry2018efficient} rotate the current gradient when the angle between the current gradient and the gradient
computed on the reference memory is obtuse. 
MEGA~\cite{guo2020improved} uses the loss of $X^{\textit{new}}$ and $X^{\textit{buf}}$ to adjust the relative importance of learning a new task and maintaining the past knowledge. OWM~\cite{zeng2019continual}, OGD~\cite{farajtabar2020orthogonal}, and AOP~\cite{guo2022adaptive} avoid CF by projecting
the gradients on the new task onto an orthogonal subspace of old tasks. 
These methods do not deal with gradient imbalance but represent a different family of CL approaches. 
}

\section{Preliminary}
\textbf{Problem description.} {Following~\cite{aljundi2019online}, we learn a sequence of tasks, 1, 2, ..., $t$, ..., in the online CIL setting.
In online CL, each incrementally arriving data batch is seen only once by the system. So the system effectively learns each task in one epoch. {For a task $t$, we denote its dataset as $ \{(x^{(t)}_k,y^{(t)}_k)\}^{n^t}_{k=1}$, where $n^t$ is the number of training samples in task $t$.} 
$L^{(t)}$ denotes the set of classes of task $t$.}

\textbf{Model architecture and training setting.} Our model $F$ consists of a feature extractor $h_\theta$ with the parameter set $\theta$ and a classifier $f_\phi$ with the parameter set $\phi$. It uses a replay method with a memory buffer $\mathcal{M}$. Whenever a small batch of new data $X^{\textit{new}}$ from the data stream is accumulated, it is trained jointly with a small batch of data $X^{\textit{buf}}$ sampled from $\mathcal{M}$ to update the model in one training iteration.
The model produces the logits $F(x;\theta,\phi)=f_\phi(h_\theta(x))$, which are used to calculate the loss or to predict in testing. Reservoir sampling is used for memory update. 



\textbf{Gradient imbalance (GI) on logits.}
We now introduce the proposed gradient-based analysis on logits and the concept of GI. 
In CL, each task is usually learned by minimizing the softmax cross-entropy loss, $\mathcal{L}_{ce}$:
\begin{equation}
\small {
\begin{aligned}
    \mathcal{L}_{ce}(o(x;\theta,\phi))&=-\sum^{|C_{\textit{seen}}|}_{i=1}l_{c_i}\log(p^{c_i}),\qquad
    p^{c_i}&=\frac{e^{o_{c_i}}}{\sum^{|C_{\textit{seen}}|}_{s=1}e^{o_{c_s}}}
\end{aligned}
}
\end{equation}
\normalsize where $|C_{\textit{seen}}|$ is the number of classes that the model has seen, {$l_{c_i}\in \{0,1\}$} is the one-hot label of class ${c_i}$, and $o(x;\theta,\phi)=[o_{c_1},o_{c_2},...,o_{c_{|C_{\textit{seen}}|}}]$ is the set of logit values for input $x$.
Given a training sample $x$ of class $c_i$, the gradients on logits {($c_j \ne c_i$)} 
are given by
\begin{equation}
    \frac{\partial \mathcal{L}_{ce}(o(x;\theta,\phi))}{\partial o_{c_i}} =p^{c_{i}}-1,\quad
    \frac{\partial \mathcal{L}_{ce}(o(x;\theta,\phi))}{\partial o_{c_j}} =p^{c_j}\label{eq:negative}.
\end{equation}

From Eq.~\ref{eq:negative}, we see that $x$ gives its true logit $o_{c_i}$ a \textbf{\textit{negative gradient}} and the other logits $o_{c_j}$ \textbf{\textit{positive gradients}}. As
the gradient update rule for a parameter $w$ is $w=w-lr*\nabla w$, where $lr$ is the learning rate, the negative gradient ($p^{c_{i}}-1$) results in an increase in $o_{c_i}$ for the true class $c_{i}$ and the positive gradient ($p^{c_{j}}$) results in a decrease in $o_{c_j}$ for each wrong class $c_{j}$. Thus, \textit{{the negative gradient encourages the model to output a larger probability for the true class}}, and \textit{{positive gradients help output lower probabilities for the wrong classes}}. 

{\color{black}However, in CL, as the model has no access to the training data of previous tasks when it learns a new task incrementally, all gradients on previous classes are positive during the new task training (there are no negative gradients) (\textit{\textbf{imbalance of positive and negative gradients}}). Then the model tends to output smaller probabilities on previous classes, biasing the classification towards the new classes.

\section{Gradient Imbalance (GI) in Replaying}
\label{sec.analysis}
\begin{figure*}[h]
\begin{minipage}[b]{1\textwidth} 
\includegraphics[width=1\textwidth]{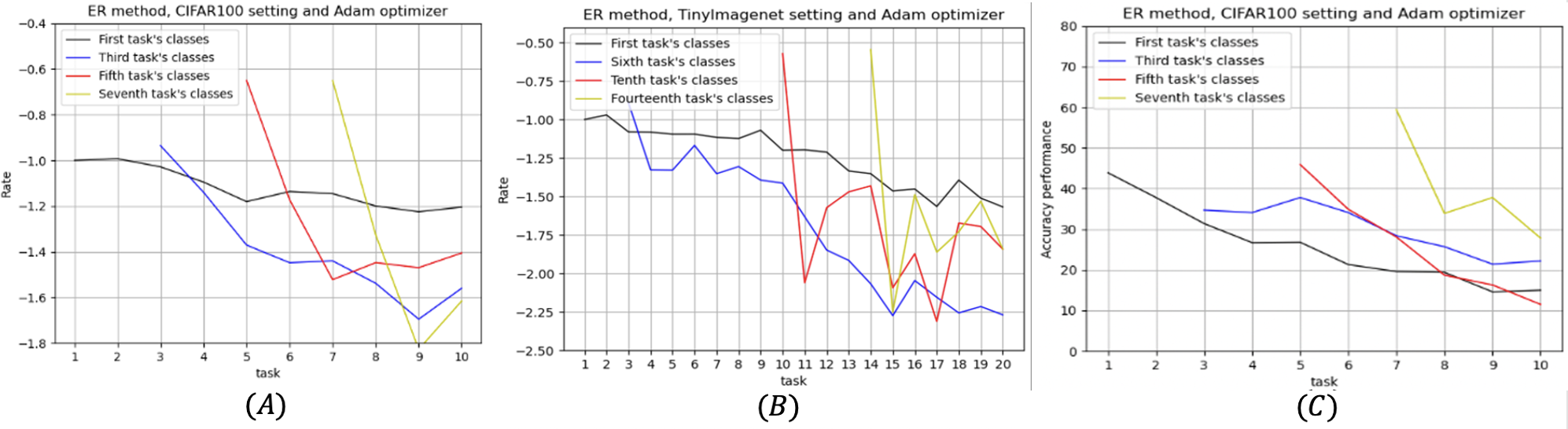} 
\end{minipage}
\caption{\textit{PN} rate (\textit{rate} in the figures) of the CIFAR100 or TinyImageNet experiments with Adam optimizers. The buffer size is 1000. We choose four different tasks in each experiment and plot their \textit{PN} rates as subsequent tasks are learned. In (C), we report their accuracy. {The plots based on the SGD optimizer are given in {\color{black}Appendix 1}, which show the same trend.}
}
\label{Fig.1}
\vspace{-2mm}
\end{figure*}
A replay-based method 
actually makes the (positive and negative) gradients more balanced. There are adjustments of both ups and downs for logits for previous classes. However, this is insufficient. We discuss two reasons: \textit{data imbalance} and \textit{CL imbalance}. We propose some metrics first.} 


\textbf{Metrics:} 
When the model is learning the new task $t$, for a seen class $c_i$ in any seen task (including the current task $t$), we define the mean of \textit{positive gradients ($p_k^{c_i}$)} that the logits of all other classes than $c_i$ have and the mean of \textit{negative gradients} ($p_k^{c_i}-1$) that class $c_i$'s logit has over {\color{black}all $n^{t,r}$ training samples (including both the data samples seen so far from the current task $t$ and the corresponding replay data used)}
for this task as 
{\small\begin{equation}
    \begin{aligned}
    P(t,c_i)&=\frac{\sum_{k=1}^{n^{t,r}}p_k^{c_i}\cdot \mathbb{I}(y_k \neq c_i) }{n^t} \\
    N(t,c_i)&=\frac{ \sum_{k=1}^{n^{t,r}}(p_k^{c_i}-1)\cdot \mathbb{I}(y_k = c_i)}{n^t}
    \end{aligned}
\end{equation}}
where $y_k$ is the label of the $k$th sample of task $t$ and $\mathbb{I}$ is the indicator function.

At the task level, for classes in any task $t'$ ($t'\leq t$), we define the mean of \textit{positive} and \textit{negative} \textit{gradients} that the logits of the class set $L^{(t')}$ receives in training task $t$ respectively as
{\small\begin{equation}
    \begin{aligned}
    P(t,L^{(t')})&=\frac{\sum_{j=1}^{|C_{\textit{seen}}|}P(t,c_j)\cdot \mathbb{I}(c_j \in L^{(t')})}{\sum_{j=1}^{|C_{\textit{seen}}|}1\cdot \mathbb{I}(c_j \in L^{(t')})}\\
    N(t,L^{(t')})&=\frac{\sum_{j=1}^{|C_{\textit{seen}}|}N(t,c_j)\cdot \mathbb{I}(c_j \in L^{(t')})}{\sum_{j=1}^{|C_{\textit{seen}}|}1\cdot \mathbb{I}(c_j \in L^{(t')})}
    \end{aligned}
\end{equation}}
\normalsize

\subsection{Gradient Imbalance due to Data Imbalance} \label{sec.imbalance1}


In a replay-based method, the memory buffer size is usually fixed and small and the batch size $N^{\textit{buf}}$ for the buffer batch $X^{\textit{buf}}$ and the batch size $N^{\textit{new}}$ for the new data batch $X^{\textit{new}}$ are also fixed. As the number of previous classes grows with more tasks learned, {the number of sampled data for each previous class $\frac{N^{\textit{buf}}}{\bigcup_{r=1}^{t-1}|L^{(r)}|}$ in $X^{\textit{buf}}$ gets smaller, but the number of samples for each new class $\frac{N^{\textit{new}}}{|L^{(t)}|}$ in $X^{\textit{new}}$ of the new task remains unchanged.}
Then we have $\frac{N^{\textit{new}}}{|L^{(t)}|} > \frac{N^{\textit{buf}}}{\bigcup_{r=1}^{t-1}|L^{(r)}|}$ and the samples from previous classes and the new data classes can become highly imbalanced. The positive gradients for the previous classes can surpass their negative gradients (in absolute values). 
{We can empirically verify this by calculating the \textit{positive-negative} (PN) \textit{gradient rate} with $t' \leq t$:
\small\begin{equation}
    \textit{PN}(t,t')=\frac{P(t,L^{(t')})}{N(t,L^{(t')})}
\end{equation}

\normalsize \textbf{Empirical Verification:} 
We introduce the setup first.

\textit{Datasets and Tasks:} We conduct our experiments using two dataset settings: (1) Split CIFAR100, where we divide the CIFAR100 dataset into 10 tasks with 10 unique classes per task, and (2) Split TinyImagenet, where we divide the TinyImageNet dataset into 20 tasks with 10 different classes per task. For online CL, we run each task in one epoch. %

\textit{Model, Optimizer and Batch Size:} 
We use the full size ResNet-18 to perform our experiments. {To give a more general analysis, we run with both SGD and Adam optimizers. For learning rate, we follow~\cite{aljundi2019online} and set it as 0.1 for the SGD optimizer. To ensure a good performance, we search and set the learning rate as 0.001 for the Adam optimizer.} For batch size, we also follow~\cite{aljundi2019online} and set it to 10 for $X^{\textit{new}}$ and 10 for the buffer batch $X^{\textit{buf}}$, which is randomly sampled from the memory buffer. {Since ER (Experience Replay) 
is a basic replay method, we conduct our experiments using ER~\cite{chaudhry2019continual}.}

We show the PN results of the classes of a few tasks of the two datasets in Fig.~\ref{Fig.1} as more subsequent tasks are learned. We can make the following observations from~Fig. \ref{Fig.1}. See plots based on SGD in {\color{black}Appendix 1} (see footnote 2). }

\textbf{(1).} Sub-figures (A) and (B) in Fig.~\ref{Fig.1} show that when the model is trained with the new task $t$, the \textit{PN} rates for the classes of old tasks get smaller than $-1$ and tend to decrease. This indicates that positive gradients surpass negative gradients (in absolute values) for the old classes. {\color{black}The reason is that the new data classes  dominate the training, and the new data give high positive gradients to classes of the old tasks to ensure the new task data are not classified to those old classes.~As more tasks are learned, the data get even more imbalanced, resulting in lower negative gradients on the old classes from the replay data of the old task $t'$ and thus the downward trend of the curves in (A) and (B). 
}

\textbf{(2)}. Sub-figure (C) in Fig.~\ref{Fig.1} shows that the tendency of the test accuracy performance of the previous classes is similar to the tendency of their \textit{PN} rates. 

\textbf{Dealing with Data Imbalance}. We propose a sampling strategy that guarantees the same number of training samples for each class in each training iteration. Specifically, we fix $N^{\textit{new}}+N^{\textit{buf}}=20$, where $N^{\textit{new}}$ and $N^{\textit{buf}}$ are the number of samples in $X^{\textit{new}}$ and $X^{\textit{buf}}$, respectively. We use the ratio between $|L^{(t)}|$ and $\sum_{r=1}^{t}|L^{(r)}|$ to decide the number of samples from $X^{\textit{new}}$ (the rest are not used here), 
i.e.,{ $N^{\textit{new}}=max(int(20\cdot \frac{|L^{(t)}|}{\sum_{r=1}^{t}|L^{(r)}|}),1)$ where $int(\cdot)$ returns the nearest integer of a given number} and  $N^{\textit{buf}}=20-N^{\textit{new}}$. We use $N^{\textit{buf}}$ to sample previous tasks' data in the memory buffer. In this way, the number of training data for every class is approximately equal.

\begin{figure*}
\vspace{-2mm}
\centering
\begin{minipage}[b]{1\textwidth} 
\includegraphics[width=1\textwidth]{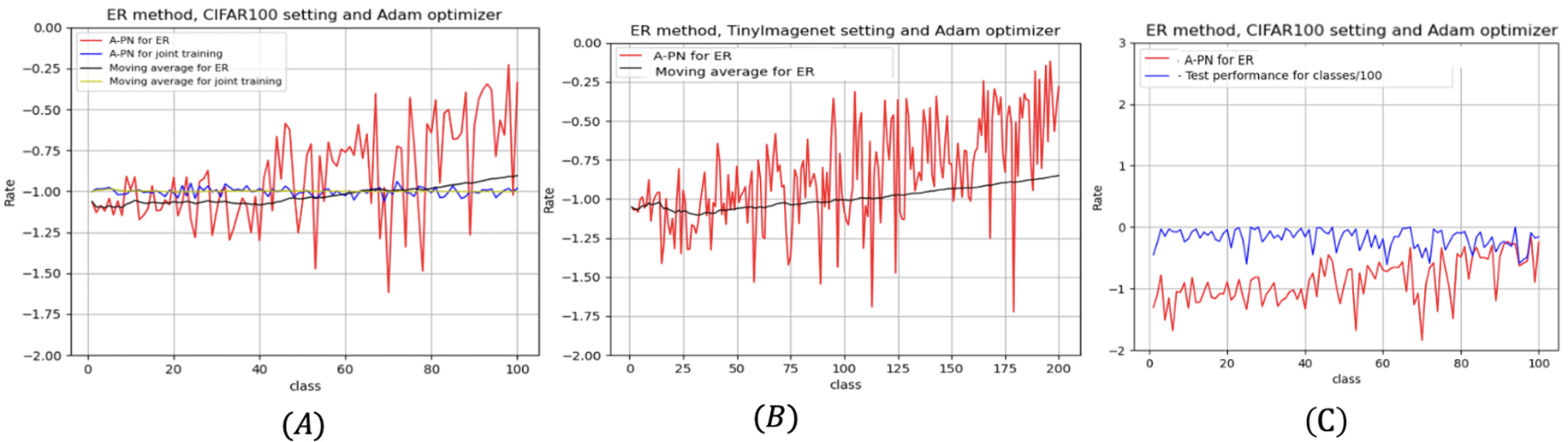} 
\end{minipage}
\vspace{-5mm}
\caption{A-PN rate (\textit{rate}) for each class ($x$-axis) after the last task is trained. {\color{black}The classes are ordered by their sequence of appearances in the original data.} The buffer size is 1000 and the optimizer is Adam. {(C) shows the test performances (see the formula in the legion) of all learned classes. The results based on the SGD optimizer are given in {\color{black}Appendix 3}.}
} 
\label{Fig.3}
\vspace{-3mm}
\end{figure*}

\subsection{Gradient Imbalance due to CL Imbalance} \label{sec.imbalance2}

{GI still occurs even after the number of samples used in each class is balanced using the technique above. Let us call the balanced data for training $X^{\textit{mix}}$, which includes both the replay data from previous tasks and new data from the current task. We now introduce the second cause of GI, \textit{CL imbalance}, which is due to incremental training in CL. 
To explore CL imbalance, we define the \textit{rate of accumulated positive and negative gradients} (A-PN rate) for {a class $c_i\in L^{(t')}$ from task $t'$ to the current task $t$ ($t' \leq t$),
\begin{equation}
\small {
    \textit{A-PN}(t',t,c_i)= \frac{\sum_{r=t'}^{t}P(r,c_i)}{\sum_{r=t'}^{t}N(r,c_i)}
    }
\label{eq:3}
\end{equation}
$\textit{A-PN}(t',t,c_i)$ gives the accumulated gradient rate of each class over the learning process to the current task $t$. 
}

\textbf{Empirical Verification:} {We plot the A-PN rate of each class in Fig.~\ref{Fig.3}(A) for the CIFAR100 dataset and in Fig.~\ref{Fig.3}(B) for the TinyImageNet dataset after the last task is learned. { 
We make the following observations from Fig.~\ref{Fig.3} (A), (B), (C) and some more details in the experiments: 

{\color{black}\textbf{(1)}. The A-NP rate accumulated is close to -1 or balanced when we jointly train all classes in the CIFAR100 dataset as a single task (the blue curve and its moving average yellow curve in Fig.~\ref{Fig.3} (A)). We also observe balanced gradients from detailed results (not shown here) for all classes when we learn the first task (no replay data). 

\textbf{(2).} We ignore the accumulative part in Eq.~\ref{eq:3} for the time being and focus on the gradient imbalance in the classes of each task alone. Based on the detailed results (not given here), we observe that as more tasks are learned, the GI for the classes of the current task being learned gets worse and worse. This explains why the classes of the last task in sub-figures (A) and (B) have the highest GI. Note, the last task has no accumulation as $t'=t$. Let us explain why. 

As an example, we consider class $c_i$ from the current task $t$. As more and more tasks are learned, the replay data from previous tasks in $X^{\textit{mix}}$ will contribute less and less positive gradients to $c_i$ because the replay data may have been trained many times in the past and are well overfitted to their own classes in the past and their probability of being classified to the new class $c_i$ is very small and hence they contribute very low positive gradients to $c_i$. {\color{black}In our experiments, we observed this and also decreased negative gradient on $c_i$.~We plot the average cross-entropy loss of the new data batch and the replay data batch when the model learns the second task of
CIFAR100 in {\color{black}Appendix 2}.} The figure shows that the average cross-entropy loss of the new data batch is much higher than that of the replay data batch in the whole training process. This is because the system has learned good features from previous tasks that makes the learning of the new class $c_i$ easier resulting in less negative gradients.~But positive gradients drop more significantly. 

We now consider the accumulation part (the summations in Eq.~\ref{eq:3}). The next section will show that A-PN is used to dynamically adjust the loss for each class in learning to balance the gradients. By right, when we learn the current task, we only need to consider the current GI situation to adjust the loss. However, considering only one case (the current situation) is risky due to random fluctuations. That is why we consider the impact on a class $c_i$ from all subsequent tasks, which gives us a more robust gradient rate for adjusting the loss in dealing with GI in learning.

\textbf{(3).} Fig.~\ref{Fig.3}(C) shows that higher A-PN rates (imbalanced gradients) result in lower test accuracy. 
The fact that the accumulated negative gradient being greater than  the accumulated positive gradients in absolute values (i.e., $\textit{A-PN} > -1$) makes the model biased towards the new classes. 

}

\section{The Proposed GSA Method}
\label{sec.proposed}
Our method GSA 
consists of two parts: (1) a new optimization framework that separately optimizes \textit{cross-task classification} and \textit{within-task classification}. (2) A \textit{gradient-based self-adaptive loss} to alleviate the gradient imbalance (CL imbalance) in our framework.

\subsection{Optimizing Within-class Classification and Cross-task Classification}
Assume that the system has seen $n-1$ previous classes $(c_1,...,c_{n-1})$ and there $m-n+1$ new classes ($c_n,...,c_m$) in the current batch $X^{\textit{new}}$, for a class, say $c_n$, in the batch of new data, we decompose its learning into two parts: (1) learning the decision boundaries between $c_n$ and the other classes in the current batch $X^{\textit{new}}$ and (2) learning the decision boundaries between $c_n$ and all previous $n-1$ classes of old tasks in $X^{\textit{buf}}$. For example, for a sample $x_{c_n}$ of new class $c_n$ from $X^{\textit{new}}$, its original cross-entropy loss is:
\begin{equation}
    \mathcal{L}_{ce}(x_{c_n})=-\log(\frac{e^{o_{c_n}}}{\sum^{m}_{s=1}e^{o_{c_s}}})
\end{equation} where $o_{c_s}$ is the logit value of $x_{c_n}$ for class $c_s$. We decompose the loss (denoted by $\mathcal{L}_{decom}(x_{c_n})$) into two parts:
\begin{equation}
    \mathcal{L}_{decom}(x_{c_n})=-\log(\frac{e^{o_{c_n}}}{\sum^{m}_{s=n}e^{o_{c_s}}})-\log(\frac{e^{o_{c_n}}}{\sum^{n-1}_{s=1}e^{o_{c_s}}+e^{o_{c_n}}})
\label{loss:decom}
\end{equation}
The first part of $\mathcal{L}_{decom}$ distinguishes $c_n$ and other new classes with respect to $x_{c_n}$. Their gradient rates are similar. The second part distinguishes $c_n$ and the previous classes with respect to $x_{c_n}$. The negative interference from new classes to previous classes is limited in the second part. 
For the relationship of loss $\mathcal{L}_{decom}$ and loss $\mathcal{L}_{ce}$, we have the following proposition:

\textbf{Proposition 1.} For a sample $x_{c_n}$ from the new class $c_n$ in $X^{\textit{new}}$, the following holds \begin{equation}
    \mathcal{L}_{decom}(x_{c_n})\geq \mathcal{L}_{ce}(x_{c_n})
\end{equation}The proof is in {\color{black}Appendix 4}. Similarly, for a sampled data point $x_{c_j}$ of class $c_j (j<n)$ from the buffer data $X^{\textit{buf}}$, we define its new loss as:
\begin{equation}
    \mathcal{L}_{decom}(x_{c_j})=-\log(\frac{e^{o_{c_j}}}{\sum^{n-1}_{s=1}e^{o_{c_s}}})-\log(\frac{e^{o_{c_j}}}{\sum^{m}_{s=n}e^{o_{c_s}}+e^{o_{c_j}}})
\label{loss:decom_buf}
\end{equation}In this loss, we put the previous classes (similar gradient rates) in the first term to maintain the established decision boundaries between previous classes and put the logit of the true class and the logits of the classes in the current task in the second term to establish the decision boundaries between class $c_j$ and the new classes. The negative transfer from the new classes to previous classes is limited to the second term. We prove that $\mathcal{L}_{decom}(x_{c_j})$ is the upper bound of $\mathcal{L}_{ce}(x_{c_j})$ in {\color{black}Appendix 4}.

\textbf{Considering data balance}. To estimate the first term in Eq.~\ref{loss:decom} for class $c_n$, we use $X^{\textit{new}}$ as the data of new classes as $X^{\textit{new}}$ is usually class-balanced.\footnote{If this is not the case, we can sample some data of the same class from the memory buffer or directly duplicate the samples to guarantee it.} 
For a similar reason, we use $X^{\textit{buf}}$ to estimate the first term in Eq.~\ref{loss:decom_buf} for the previous class $c_j$.
However, for the second term in Eq.~\ref{loss:decom}, using $X^{\textit{new}}$ causes prediction bias towards new classes as $X^{\textit{new}}$ lacks samples of previous classes. Also, using $X^{\textit{buf}}$ to estimate the second term in Eq.~\ref{loss:decom_buf} is not appropriate as the number of new classes samples in the buffer increases gradually and for a long time, the new classes are the minor classes in the buffer. To solve both problems, we use the sampling strategy proposed in the last section under ``Dealing with Data Imbalance.''
Then we obtain a mixed set of new data and old data sampled from the memory buffer, called $X^{\textit{mix}}$. $X^{\textit{mix}}$ can be regarded as samples from the uniform joint distribution {for all classes, including both the new and old classes.} Then our final optimization goal is:
\small
\begin{equation}
\begin{aligned}
 \min\limits_{F\in\mathcal{F}}-\mathop{\mathbb{E}}\limits_{x\sim X^{\textit{new}}}\log(\frac{e^{o_{y_x}}}{\sum^{m}_{s=n}e^{o_{c_s}}})-\mathop{\mathbb{E}}\limits_{x\sim X^{\textit{buf}}}\log(\frac{e^{o_{y_x}}}{\sum^{n-1}_{s=1}e^{o_{c_s}}})\\-\mathop{\mathbb{E}}\limits_{x\sim X^{\textit{mix}}}({\mathbb{I}(y_x\cap S(X^{\textit{new}})\neq\emptyset)}\cdot\log(\frac{e^{o_{y_x}}}{\sum^{n-1}_{s=1}e^{o_{c_s}}+e^{o_{y_x}}})+\\{\mathbb{I}(y_x\cap S(X^{\textit{new}})=\emptyset)}\cdot\log(\frac{e^{o_{y_x}}}{\sum^{m}_{s=n}e^{o_{c_s}}+e^{o_{y_x}}}))
 \label{loss_op}
\end{aligned}
\end{equation} \normalsize where $y_x$ is the label of sample $x$ and $S(X^{\textit{new}})$ is the set of classes appeared in the $X^{\textit{new}}$ and $\mathbb{I}$ is the indicator function and $\mathcal{F}$ is the function space. The first term and second terms focus on the within-task classification goal and the last term focus on the cross-task classification goal.

\subsection{Self-Adaptive Loss for CL Imbalance}
 Based on the analysis in Section~\ref{sec.analysis}, we know that even though the data balance is guaranteed, GI still exists as the gradient rate is different for different classes, which is caused by CL imbalance. We note that CL imbalance (measured by the gradient rate of each class) not only occurs between new and previous classes but also occurs among previous classes. To mitigate it, we propose \textit{GSA-CE} (Gradient-based Self-Adaptive CE) loss $\mathcal{L}_{\textit{GSA-CE}}$. The loss function is \textit{dynamically adjusted} based on the gradient rates, which change as more tasks and batches are learned. 
{\color{black}Let $\{x_k,y_k\}_{k=1}^{N^{\textit{mix}}}$ be the samples in $X^{\textit{mix}}$, their $\mathcal{L}_{\textit{GSA-CE}}$ is:
\small
\begin{equation}
\begin{aligned}
 \mathcal{L}_{\textit{GSA-CE}}(\{x_k,y_k\}_{k=1}^{N^{\textit{mix}}}=-\frac{1}{N^{\textit{mix}}}\sum_{k=1}^{N^{\textit{mix}}}w_{y_k}\cdot({\mathbb{I}(y_k\cap S(X^{\textit{new}})\neq\emptyset)}\cdot\\\log(\frac{e^{o_{y_k}}}{\sum^{n-1}_{s=1}e^{o_{c_s}}+e^{o_{y_k}}})+{\mathbb{I}(y_k\cap S(X^{\textit{new}})=\emptyset)}\cdot\\\log(\frac{e^{o_{y_k}}}{\sum^{m}_{s=n}v_{c_s}e^{o_{c_s}}+e^{o_{y_k}}}))
 \label{eq:mix}
\end{aligned}
\end{equation}
\normalsize
where }
\begin{equation}
\small
    w_{y_k}=\frac{2}{1+e^{\textit{A-PN}(t',t,y_k)}}, \quad v_{c_s}=\frac{1}{-\textit{PN}(t,c_s)}
\end{equation}
\normalsize
where $t'$ is the task in which class $c_s$ appears. 
When the model is training task $t$, 
{we incrementally update the $\textit{A-PN}(t',t,y_k)$ and $\textit{PN}(t,y_k)$ by adding the new gradients into the current sums of all previous gradients. 
 $\mathcal{L}_{\textit{GSA-CE}}$} has three advantages: 

\textbf{(1).}~When the new task arrives, we do not need to assume that we have the entire training set for calculating some statistics or an exemplar set like CCIL~\cite{mittal2021essentials}. Our method is thus suitable for online CL.

\textbf{(2).}~{The model sets the values of $w_{y_k}$ and $v_{c_s}$ based on \textit{A-PN} rates and \textit{PN} rates automatically as they vary dynamically with the current gradient imbalance situation.} 

\textbf{(3).}~\textit{A-PN} and \textit{PN} rates are class-based metrics that reflect the discriminative power of the model for all seen classes. Our loss then gives class-based weights for samples from different classes. The new loss thus automatically adjust the loss weight $w_{y_k}$ to balance the accumulated negative gradient and the accumulated positive gradient for each class. We justify the new loss in {\color{black}Appendix 5}.

\textbf{(4)} No new hyperparameter is introduced in the process. 

\vspace{+2mm}
Similarly, as the gradient rate imbalance occurs also within previous classes, for samples $\{x_k,y_k\}_{k=1}^{N^{\textit{buf}}}$ from $X^{\textit{buf}}$, their $\mathcal{L}_{\textit{GSA-CE}}$ loss is:
\small
\begin{equation}
\begin{aligned}
&\mathcal{L}_{\textit{GSA-CE}}(\{x_k,y_k\}_{k=1}^{N^{\textit{buf}}})=-\frac{\sum_{k=1}^{N^\textit{buf}}w_{y_k}\log(\frac{e^{o_{y_k}}}{\sum^{n-1}_{s=1}e^{o_{c_s}}})}{N^{\textit{buf}}}
\label{eq:buf}
\end{aligned}
\end{equation}
\normalsize
Considering the gradient imbalance, we replace the second term and the third term in Eq.~\ref{loss_op} with our $\mathcal{L}_{\textit{GSA-CE}}$ losses Eq.~\ref{eq:buf} and Eq.~\ref{eq:mix} in the expectation form respectively. The final optimization objective in Eq.~\ref{loss_op} is turned into:
\small
\begin{equation}
\begin{aligned}
 \min\limits_{F\in\mathcal{F}}-\mathop{\mathbb{E}}\limits_{x\sim X^{\textit{new}}}\log(\frac{e^{o_{y_x}}}{\sum^{m}_{s=n}e^{o_{c_s}}})-\mathop{\mathbb{E}}\limits_{x\sim X^{\textit{buf}}} w_{y_x}\cdot\log(\frac{e^{o_{y_x}}}{\sum^{n-1}_{s=1}e^{o_{c_s}}})\\-\mathop{\mathbb{E}}\limits_{x\sim X^{\textit{mix}}}w_{y_x}\cdot({\mathbb{I}(y_x\cap S(X^{\textit{new}})\neq\emptyset)}\cdot\log(\frac{e^{o_{y_x}}}{\sum^{n-1}_{s=1}e^{o_{c_s}}+e^{o_{y_x}}})\\+{\mathbb{I}(y_x\cap S(X^{\textit{new}})=\emptyset)}\cdot\log(\frac{e^{o_{y_x}}}{\sum^{m}_{s=n}v_{c_s}e^{o_{c_s}}+e^{o_{y_x}}}))
 \label{eq:final}
\end{aligned}
\end{equation}
\normalsize

}}}

\section{Experiments} 
\label{sec.experiments}
\begin{table*}[h!]
\centering
\renewcommand{\arraystretch}{1.1}
\footnotesize
\scalebox{0.80}{
\renewcommand\tabcolsep{0.6mm}
\begin{tabular}{c|ccc|ccc|ccc|ccc}
\hline
{\bf Method}&{}&{MNIST (5 tasks)}&{}&{}&{CIFAR10 (5 tasks)}&{}&{}&{CIFAR100 (10 tasks)}&{}&{}&{TinyImageNet (100 tasks)}&{}\\
\hline
\textit{M}&{\textit{M}=0.1\textit{k}}&{\textit{M}=0.5\textit{k}}&{\textit{M}=1\textit{k}}&{\textit{M}=0.2\textit{k}}&{\textit{M}=0.5\textit{k}}&{\textit{M}=1\textit{k}}&{\textit{M}=1\textit{k}}&{\textit{M}=2\textit{k}}&{\textit{M}=5\textit{k}}&{\textit{M}=2\textit{k}}&{\textit{M}=4\textit{k}}&{\textit{M}=10\textit{k}}
\\
\hline
AGEM~\cite{chaudhry2018efficient} &56.9{\scriptsize$\pm$5.2}&57.7{\scriptsize$\pm$8.8}&61.6{\scriptsize$\pm$3.2}&22.7{\scriptsize$\pm$1.8}&22.7{\scriptsize$\pm$1.9}&22.6{\scriptsize$\pm$0.7}&5.8{\scriptsize$\pm$0.2}&6.5{\scriptsize$\pm$0.3}&5.8{\scriptsize$\pm$0.2}&0.9{\scriptsize$\pm$0.1}&2.1{\scriptsize$\pm$0.1}&3.9{\scriptsize$\pm$0.2}
\\
\hline
ER~\cite{chaudhry2019continual}& 78.7{\scriptsize$\pm$0.4}&88.0{\scriptsize$\pm$0.2} & 90.3{\scriptsize$\pm$0.1} & 49.7{\scriptsize$\pm$0.6}&55.2{\scriptsize$\pm$0.6}&59.3{\scriptsize$\pm$0.2}&15.7{\scriptsize$\pm$0.3}&22.4{\scriptsize$\pm$0.5}&29.5{\scriptsize$\pm$0.9}&4.7{\scriptsize$\pm$0.5}&10.1{\scriptsize$\pm$0.7}&11.7{\scriptsize$\pm$0.2} 
\\
\hline
MIR~\cite{aljundi2019online}  &79.0{\scriptsize$\pm$0.5}&88.3{\scriptsize$\pm$0.1}&91.3{\scriptsize$\pm$1.9}&37.3{\scriptsize$\pm$0.3}&40.0{\scriptsize$\pm$0.6}&41.0{\scriptsize$\pm$0.6}&15.7{\scriptsize$\pm$0.2}&19.1{\scriptsize$\pm$0.1}&24.1{\scriptsize$\pm$0.2}&6.1{\scriptsize$\pm$0.5}&11.7{\scriptsize$\pm$0.2}&13.5{\scriptsize$\pm$0.2}
\\
\hline
GSS~\cite{aljundi2019gradient} &70.4{\scriptsize$\pm$1.5}&80.7{\scriptsize$\pm$5.8}&87.5{\scriptsize$\pm$5.9}&26.9{\scriptsize$\pm$1.2}&30.7{\scriptsize$\pm$1.3}&40.1{\scriptsize$\pm$1.4}&11.1{\scriptsize$\pm$0.2}&13.3{\scriptsize$\pm$0.5}&17.4{\scriptsize$\pm$0.1}&3.3{\scriptsize$\pm$0.5}&10.0{\scriptsize$\pm$0.2}&10.5{\scriptsize$\pm$0.2}
\\
\hline
ASER~\cite{shim2021online}  &61.6{\scriptsize$\pm$2.1}&71.0{\scriptsize$\pm$0.6}&82.1{\scriptsize$\pm$5.9}&27.8{\scriptsize$\pm$1.0}&36.2{\scriptsize$\pm$1.2}&44.7{\scriptsize$\pm$1.2}&16.4{\scriptsize$\pm$0.3}&12.2{\scriptsize$\pm$1.9}&27.1{\scriptsize$\pm$0.3}&5.3{\scriptsize$\pm$0.3}&8.2{\scriptsize$\pm$0.2}&10.3{\scriptsize$\pm$0.4}
\\
\hline
Rainbow-rt~\cite{bang2021rainbow}
&89.1{\scriptsize$\pm$0.3}&92.1{\scriptsize$\pm$0.1}&95.0{\scriptsize$\pm$0.3}&45.2{\scriptsize$\pm$0.4}&50.6{\scriptsize$\pm$0.3}&51.5{\scriptsize$\pm$0.8}&15.4{\scriptsize$\pm$0.5}&15.9{\scriptsize$\pm$0.3}&20.4{\scriptsize$\pm$0.3}&6.6{\scriptsize$\pm$0.3}&10.1{\scriptsize$\pm$0.3}&13.1{\scriptsize$\pm$0.5}
\\
\hline
ER-AML~\cite{caccia2021reducing}
&76.5{\scriptsize$\pm$0.1}&86.0{\scriptsize$\pm$0.2}&91.5{\scriptsize$\pm$0.1}&40.5{\scriptsize$\pm$0.7}&48.7{\scriptsize$\pm$0.5}&50.1{\scriptsize$\pm$0.4}&16.1{\scriptsize$\pm$0.4}&17.6{\scriptsize$\pm$0.5}&22.6{\scriptsize$\pm$0.1}&5.4{\scriptsize$\pm$0.2}&7.1{\scriptsize$\pm$0.5}&10.1{\scriptsize$\pm$0.4}\\
\hline
GDumb~\cite{prabhu2020gdumb} &81.2{\scriptsize$\pm$0.5}&91.0{\scriptsize$\pm$0.2}&94.5{\scriptsize$\pm$0.1}&35.9{\scriptsize$\pm$1.1}&50.7{\scriptsize$\pm$0.7}&63.5{\scriptsize$\pm$0.5}&17.1{\scriptsize$\pm$0.4}&25.1{\scriptsize$\pm$0.2}&38.6{\scriptsize$\pm$0.5}&12.6{\scriptsize$\pm$0.1}&12.7{\scriptsize$\pm$0.3}&15.7{\scriptsize$\pm$0.2}\\
\hline
SCR~\cite{Mai_2021_CVPR} &86.2{\scriptsize$\pm$0.5}&\text{92.8}{\scriptsize$\pm$0.3}&94.6{\scriptsize$\pm$0.1}&47.2{\scriptsize$\pm$1.7}&58.2{\scriptsize$\pm$0.5}&64.1{\scriptsize$\pm$1.2}&26.5{\scriptsize$\pm$0.2}&31.6{\scriptsize$\pm$0.5}&36.5{\scriptsize$\pm$0.2}&10.6{\scriptsize$\pm$1.1}&17.2{\scriptsize$\pm$0.1}&20.4{\scriptsize$\pm$1.1}\\
\hline
\hline
DER++~\cite{buzzega2020dark}  &74.4{\scriptsize$\pm$1.1}&91.5{\scriptsize$\pm$0.2}&92.1{\scriptsize$\pm$0.2}&44.2{\scriptsize$\pm$1.1}&47.9{\scriptsize$\pm$1.5}&54.7{\scriptsize$\pm$2.2}&15.3{\scriptsize$\pm$0.2}&19.7{\scriptsize$\pm$1.5}&27.0{\scriptsize$\pm$0.7}&4.5{\scriptsize$\pm$0.3}&10.1{\scriptsize$\pm$0.3}&17.6{\scriptsize$\pm$0.5}\\
\hline
IL2A~\cite{zhu2021class}&90.2{\scriptsize$\pm$0.1}&92.7{\scriptsize$\pm$0.1}&93.9{\scriptsize$\pm$0.1}&54.7{\scriptsize$\pm$0.5}&56.0{\scriptsize$\pm$0.4}&58.2{\scriptsize$\pm$1.2}&18.2{\scriptsize$\pm$1.2}&19.7{\scriptsize$\pm$0.5}&22.4{\scriptsize$\pm$0.2}&5.5{\scriptsize$\pm$0.7}&8.1{\scriptsize$\pm$1.2}&11.6{\scriptsize$\pm$0.4}\\
\hline
Co$^2$L~\cite{Cha_2021_ICCV}&83.1{\scriptsize$\pm$0.1}&91.5{\scriptsize$\pm$0.1}&94.7{\scriptsize$\pm$0.1}&42.1{\scriptsize$\pm$1.2}&51.0{\scriptsize$\pm$0.7}&58.8{\scriptsize$\pm$0.4}&17.1{\scriptsize$\pm$0.4}&24.2{\scriptsize$\pm$0.2}&32.2{\scriptsize$\pm$0.5}&10.1{\scriptsize$\pm$0.2}&15.8{\scriptsize$\pm$0.4}&22.5{\scriptsize$\pm$1.2}\\
\hline
LUCIR~\cite{hou2019learning}  &73.2{\scriptsize$\pm$0.1}&\text{87.2}{\scriptsize$\pm$0.2}&90.1{\scriptsize$\pm$0.1}&27.9{\scriptsize$\pm$1.2}&33.5{\scriptsize$\pm$0.7}&37.8{\scriptsize$\pm$0.5}&8.6{\scriptsize$\pm$1.3}&19.5{\scriptsize$\pm$0.7}&16.9{\scriptsize$\pm$0.5}&7.6{\scriptsize$\pm$0.5}&9.6{\scriptsize$\pm$0.7}&12.5{\scriptsize$\pm$0.7}\\
\hline
CCIL~\cite{mittal2021essentials} &86.4{\scriptsize$\pm$0.1}&\text{92.8}{\scriptsize$\pm$0.2}&94.0{\scriptsize$\pm$0.1}&50.5{\scriptsize$\pm$0.2}&55.3{\scriptsize$\pm$0.5}&59.8{\scriptsize$\pm$0.3}&18.5{\scriptsize$\pm$0.3}&19.1{\scriptsize$\pm$0.4}&20.5{\scriptsize$\pm$0.3}&5.6{\scriptsize$\pm$0.9}&7.0{\scriptsize$\pm$0.5}&15.2{\scriptsize$\pm$0.5}\\
\hline
BiC~\cite{wu2019large} &90.4{\scriptsize$\pm$0.1}&\text{93.0}{\scriptsize$\pm$0.2}&94.8{\scriptsize$\pm$0.1}&48.2{\scriptsize$\pm$0.7}&57.5{\scriptsize$\pm$1.4}&63.8{\scriptsize$\pm$0.2}&21.2{\scriptsize$\pm$0.3}&36.1{\scriptsize$\pm$1.3}&42.5{\scriptsize$\pm$1.2}&10.2{\scriptsize$\pm$0.9}&18.9{\scriptsize$\pm$0.3}&25.2{\scriptsize$\pm$0.6}\\
\hline
SSIL~\cite{ahn2021ss}
&88.2{\scriptsize$\pm$0.1}&\text{93.0}{\scriptsize$\pm$0.2}&95.1{\scriptsize$\pm$0.1}&49.5{\scriptsize$\pm$0.2}&59.2{\scriptsize$\pm$0.4}&64.0{\scriptsize$\pm$0.5}&26.0{\scriptsize$\pm$0.1}&33.1{\scriptsize$\pm$0.5}&39.5{\scriptsize$\pm$0.4}&9.6{\scriptsize$\pm$0.7}&15.2{\scriptsize$\pm$1.5}&21.1{\scriptsize$\pm$0.1}\\
\hline
\hline
GSA &{\textbf{91.4}}{\scriptsize$\pm$0.1} &{\textbf{93.2}}{\scriptsize$\pm$0.1}  &{\textbf{96.5}}{\scriptsize$\pm$0.1}&\textbf{58.0}{\scriptsize$\pm$0.4} &{\textbf{64.6}}{\scriptsize$\pm$0.2}&{\textbf{69.1}}{\scriptsize$\pm$0.3}&\textbf{31.4}{\scriptsize$\pm$0.2} &\textbf{39.7}{\scriptsize$\pm$0.6}&\textbf{49.7}{\scriptsize$\pm$0.2}&\textbf{18.4}{\scriptsize$\pm$0.4} &\textbf{26.0}{\scriptsize$\pm$0.2}&\textbf{33.2}{\scriptsize$\pm$0.4}
\\
\hline
\end{tabular}
}
\caption{Accuracy on the four experiment datasets 
with different memory buffer sizes \textit{M}. All values are averages of 15 runs.}
\vspace{-3mm}
\label{tab:accresults}
\end{table*}

\begin{table*}[h!]
\centering
\renewcommand{\arraystretch}{1.1}
\footnotesize
\scalebox{0.80}{
\renewcommand\tabcolsep{0.6mm}
\begin{tabular}{c|ccc|ccc|ccc|ccc}
\hline
{\bf Method}&{}&{MNIST (5 tasks)}&{}&{}&{CIFAR10 (5 tasks)}&{}&{}&{CIFAR100 (10 tasks)}&{}&{}&{TinyImageNet (100 tasks)}&{}\\
\hline
\textit{M}&{\textit{M}=0.1\textit{k}}&{\textit{M}=0.5\textit{k}}&{\textit{M}=1\textit{k}}&{\textit{M}=0.2\textit{k}}&{\textit{M}=0.5\textit{k}}&{\textit{M}=1\textit{k}}&{\textit{M}=1\textit{k}}&{\textit{M}=2\textit{k}}&{\textit{M}=5\textit{k}}&{\textit{M}=2\textit{k}}&{\textit{M}=4\textit{k}}&{\textit{M}=10\textit{k}}
\\
\hline
AGEM &32.5{\scriptsize$\pm$5.9}&30.1{\scriptsize$\pm$4.2}&32.0{\scriptsize$\pm$2.9}&36.1{\scriptsize$\pm$3.8}&43.2{\scriptsize$\pm$4.3}&48.1{\scriptsize$\pm$3.4}&78.6{\scriptsize$\pm$2.1}&77.5{\scriptsize$\pm$1.3}&78.3{\scriptsize$\pm$1.2}&73.9{\scriptsize$\pm$0.2}&78.9{\scriptsize$\pm$0.2}&74.1{\scriptsize$\pm$0.3}
\\
\hline
ER& 22.7{\scriptsize$\pm$0.5}&9.7{\scriptsize$\pm$0.4} & 6.7{\scriptsize$\pm$0.5} & 42.0{\scriptsize$\pm$0.3}&26.7{\scriptsize$\pm$0.7}&20.7{\scriptsize$\pm$0.7}&65.1{\scriptsize$\pm$1.3}&59.3{\scriptsize$\pm$0.9}&60.0{\scriptsize$\pm$1.6}&68.2{\scriptsize$\pm$2.8}&66.2{\scriptsize$\pm$0.8}&67.2{\scriptsize$\pm$0.2} \\
\hline
MIR~\cite{aljundi2019online}  &22.3{\scriptsize$\pm$0.5}&9.0{\scriptsize$\pm$0.5}&5.7{\scriptsize$\pm$0.9}&40.0{\scriptsize$\pm$1.6}&25.9{\scriptsize$\pm$0.7}&24.5{\scriptsize$\pm$0.5}&24.5{\scriptsize$\pm$0.3}&21.4{\scriptsize$\pm$0.3}&21.0{\scriptsize$\pm$0.1}&61.1{\scriptsize$\pm$3.2}&60.9{\scriptsize$\pm$0.3}&59.5{\scriptsize$\pm$0.3}
\\
\hline
GSS &26.1{\scriptsize$\pm$2.2}&17.8{\scriptsize$\pm$5.22}&10.5{\scriptsize$\pm$6.7}&75.5{\scriptsize$\pm$1.5}&65.9{\scriptsize$\pm$1.6}&54.9{\scriptsize$\pm$2.0}&73.4{\scriptsize$\pm$4.2}&69.3{\scriptsize$\pm$3.1}&70.9{\scriptsize$\pm$2.9}&72.8{\scriptsize$\pm$1.2}&72.6{\scriptsize$\pm$0.4}&71.5{\scriptsize$\pm$0.2}
\\
\hline
ASER~\cite{shim2021online}&33.8{\scriptsize$\pm$1.1}&24.8{\scriptsize$\pm$0.5}&13.8{\scriptsize$\pm$0.4}&71.1{\scriptsize$\pm$1.8}&59.1{\scriptsize$\pm$1.5}&50.4{\scriptsize$\pm$1.5}&25.0{\scriptsize$\pm$0.2}&12.2{\scriptsize$\pm$1.9}&13.2{\scriptsize$\pm$0.1}&65.7{\scriptsize$\pm$0.7}&64.2{\scriptsize$\pm$0.2}&62.2{\scriptsize$\pm$0.1}
\\
\hline
Rainbow-rt&10.1{\scriptsize$\pm$0.1}&4.7{\scriptsize$\pm$0.4}&2.4{\scriptsize$\pm$0.5}&20.4{\scriptsize$\pm$0.3}&18.1{\scriptsize$\pm$0.4}&15.3{\scriptsize$\pm$0.8}&25.5{\scriptsize$\pm$0.5}&19.3{\scriptsize$\pm$0.4}&13.3{\scriptsize$\pm$0.5}&25.5{\scriptsize$\pm$0.6}&23.2{\scriptsize$\pm$0.3}&20.0{\scriptsize$\pm$0.4}
\\
\hline
ER-AML&23.1{\scriptsize$\pm$0.1}&9.7{\scriptsize$\pm$0.4}&6.4{\scriptsize$\pm$0.5}&50.9{\scriptsize$\pm$0.3}&40.1{\scriptsize$\pm$0.5}&34.2{\scriptsize$\pm$0.8}&51.5{\scriptsize$\pm$0.8}&49.2{\scriptsize$\pm$0.5}&38.7{\scriptsize$\pm$0.6}&47.4{\scriptsize$\pm$0.5}&43.2{\scriptsize$\pm$0.3}&41.0{\scriptsize$\pm$0.5}
\\
\hline
GDumb &10.3{\scriptsize$\pm$0.1}&6.2{\scriptsize$\pm$0.1}&4.8{\scriptsize$\pm$0.2}&26.5{\scriptsize$\pm$0.5}&24.5{\scriptsize$\pm$0.2}&18.9{\scriptsize$\pm$0.4}&16.7{\scriptsize$\pm$0.5}&17.6{\scriptsize$\pm$0.2}&16.8{\scriptsize$\pm$0.4}&15.9{\scriptsize$\pm$0.5}&14.6{\scriptsize$\pm$0.3}&11.7{\scriptsize$\pm$0.2}
\\
\hline
SCR &10.7{\scriptsize$\pm$0.1}&4.7{\scriptsize$\pm$0.1}&4.0{\scriptsize$\pm$0.2}&41.3{\scriptsize$\pm$0.1}&31.5{\scriptsize$\pm$0.2}&24.7{\scriptsize$\pm$0.4}&17.5{\scriptsize$\pm$0.2}&11.6{\scriptsize$\pm$0.5}&5.6{\scriptsize$\pm$0.4}&19.4{\scriptsize$\pm$0.3}&15.4{\scriptsize$\pm$0.3}&14.9{\scriptsize$\pm$0.7}
\\
\hline
\hline
DER++~\cite{buzzega2020dark}  &25.0{\scriptsize$\pm$0.3}&7.3{\scriptsize$\pm$0.3}&6.6{\scriptsize$\pm$1.2}&30.1{\scriptsize$\pm$0.8}&31.8{\scriptsize$\pm$2.5}&18.7{\scriptsize$\pm$3.4}&43.4{\scriptsize$\pm$0.2}&44.0{\scriptsize$\pm$1.9}&25.8{\scriptsize$\pm$3.5}&67.2{\scriptsize$\pm$1.7}&63.6{\scriptsize$\pm$0.3}&55.2{\scriptsize$\pm$0.7}
\\
\hline
IL2A~\cite{zhu2021class}&8.7{\scriptsize$\pm$0.1}&7.2{\scriptsize$\pm$0.1}&4.1{\scriptsize$\pm$0.1}&36.0{\scriptsize$\pm$0.2}&32.1{\scriptsize$\pm$0.4}&29.1{\scriptsize$\pm$0.4}&24.6{\scriptsize$\pm$0.6}&12.5{\scriptsize$\pm$0.7}&20.0{\scriptsize$\pm$0.5}&65.5{\scriptsize$\pm$0.7}&60.1{\scriptsize$\pm$0.5}&57.6{\scriptsize$\pm$1.1}\\
\hline
Co$^2$L~\cite{Cha_2021_ICCV}&14.7{\scriptsize$\pm$0.2}&7.1{\scriptsize$\pm$0.1}&3.1{\scriptsize$\pm$0.1}&32.0{\scriptsize$\pm$0.1}&21.0{\scriptsize$\pm$0.3}&16.9{\scriptsize$\pm$0.2}&16.9{\scriptsize$\pm$0.4}&16.6{\scriptsize$\pm$0.6}&9.9{\scriptsize$\pm$0.7}&60.5{\scriptsize$\pm$0.5}&52.5{\scriptsize$\pm$0.9}&42.5{\scriptsize$\pm$0.8}\\
\hline
LUCIR &20.3{\scriptsize$\pm$0.1}&8.5{\scriptsize$\pm$0.1}&7.8{\scriptsize$\pm$0.1}&63.5{\scriptsize$\pm$0.5}&55.3{\scriptsize$\pm$0.2}&46.5{\scriptsize$\pm$1.2}&60.0{\scriptsize$\pm$0.1}&47.5{\scriptsize$\pm$0.9}&44.3{\scriptsize$\pm$0.7}&46.4{\scriptsize$\pm$0.7}&42.2{\scriptsize$\pm$0.9}&37.6{\scriptsize$\pm$0.7}
\\
\hline
CCIL &14.1{\scriptsize$\pm$0.1}&7.7{\scriptsize$\pm$0.1}&4.8{\scriptsize$\pm$0.1}&18.6{\scriptsize$\pm$0.1}&16.5{\scriptsize$\pm$0.4}&12.5{\scriptsize$\pm$0.8}&16.7{\scriptsize$\pm$0.5}&16.1{\scriptsize$\pm$0.3}&17.5{\scriptsize$\pm$0.2}&59.4{\scriptsize$\pm$0.3}&56.2{\scriptsize$\pm$1.3}&48.9{\scriptsize$\pm$0.6}
\\
\hline
BiC &11.1{\scriptsize$\pm$0.1}&3.7{\scriptsize$\pm$0.1}&2.5{\scriptsize$\pm$0.1}&35.4{\scriptsize$\pm$0.7}&25.3{\scriptsize$\pm$0.4}&14.5{\scriptsize$\pm$0.7}&40.2{\scriptsize$\pm$0.4}&30.9{\scriptsize$\pm$0.7}&18.7{\scriptsize$\pm$0.5}&43.5{\scriptsize$\pm$0.5}&32.9{\scriptsize$\pm$0.5}&24.9{\scriptsize$\pm$0.4}
\\
\hline
SSIL &11.3{\scriptsize$\pm$0.1} &2.7{\scriptsize$\pm$0.1}&2.8{\scriptsize$\pm$0.1}&36.0{\scriptsize$\pm$0.7}&29.6{\scriptsize$\pm$0.4}&13.5{\scriptsize$\pm$0.4}&40.1{\scriptsize$\pm$0.5}&33.9{\scriptsize$\pm$1.2}&21.7{\scriptsize$\pm$0.8}&44.4{\scriptsize$\pm$0.7}&36.6{\scriptsize$\pm$0.7}&29.0{\scriptsize$\pm$0.7}
\\
\hline
\hline
GSA &8.1{\scriptsize$\pm$0.1} &2.5{\scriptsize$\pm$0.1} &1.4{\scriptsize$\pm$0.1} &23.5{\scriptsize$\pm$0.2} &12.6{\scriptsize$\pm$0.4} &10.0{\scriptsize$\pm$0.3}
&33.2{\scriptsize$\pm$0.6} &22.8{\scriptsize$\pm$0.4} &8.7{\scriptsize$\pm$0.3}&35.5{\scriptsize$\pm$0.3} &25.8{\scriptsize$\pm$0.4} &16.9{\scriptsize$\pm$0.6}
\\
\hline
\end{tabular}
}
 \caption{Average forgetting rate. All numbers are the averages of 15 runs. 
 }
\label{tab:forgettingrate}
\vspace{-3mm}
\end{table*}

\begin{table}[t!]
\centering
\renewcommand{\arraystretch}{1.1}
\footnotesize
\scalebox{0.80}{
\renewcommand\tabcolsep{0.6mm}
\begin{tabular}{c|c|c|c||c|c}
\hline
{\bf Dataset}&{no new loss
}%
&
{no previous loss 
}&{no cross loss
}
&{no $X^{\textit{mix}}$}&{no CL imbalance}\\
CIFAR100 &30.1{\scriptsize$\pm$0.6}&12.7{\scriptsize$\pm$0.2}&29.9{\scriptsize$\pm$0.4}&30.1{\scriptsize$\pm$0.1}&29.7{\scriptsize$\pm$0.3}
\\
\hline
TinyImageNet &17.5{\scriptsize$\pm$0.2}&4.7{\scriptsize$\pm$0.5}&17.2{\scriptsize$\pm$0.4}&17.5{\scriptsize$\pm$0.5}&16.9{\scriptsize$\pm$0.2}
\\
\hline
\end{tabular}
}
\vspace{-2.5mm}
 \caption{Ablation accuracy - average of 5 runs. Memory size \textit{M} is 1k for CIFAR100 and 2k for TinyImageNet.}
\label{tab:ablation}
 \vspace{-7.5mm}
\end{table}

\textbf{Evaluation data.} Four image classification datasets are used. 
1) \textbf{MNIST}~\cite{leCun1998mnist}  has 10 classes with 60,000/10,000 training/test examples. We created 5 disjoint tasks with 2 classes per task. 
2) \textbf{CIFAR10}~\cite{Krizhevsky2009learning} has 10 classes with 50,000/10,000 training/test samples. We created 5 disjoint tasks with 2 classes per task.  
3) \textbf{CIFAR100}~\cite{Krizhevsky2009learning} has 100 classes with 50,000/10,000 training/test samples. 10 disjoint tasks are created with 10 classes per task. 
4) \textbf{TinyImageNet}~\cite{le2015tiny} 
has 200 classes. We created 100 disjoint tasks with 2 classes per task. 
Each class has 500 training examples and 50 test examples. 

\textbf{Compared Baselines.} GSA\footnote{Code and Appendix: https://github.com/gydpku/GSA} is compared with 16 recent baselines: 9 online CL baselines, \textbf{AGEM}, \textbf{ER}, \textbf{MIR}, \textbf{GSS}, \textbf{ASER}, \textbf{RainBow-rt}, \textbf{ER-AML}, \textbf{GDumb}, and
\textbf{SCR} 
and 7 offline CL baselines, \textbf{DER++}, \textbf{IL2A}, \textbf{Co$^2$L}, \textbf{LUCIR}, \textbf{CCIL}, \textbf{BiC} and \textbf{SSIL}, as they deal with data/class imbalance and can be run in the online CL mode, i.e., training in one epoch without requiring the full data of each task to be available when the task arrives. The citations of these systems are given in Table~\ref{tab:accresults} associated with their results. {\color{black} Note that {Rainbow}~\cite{bang2021rainbow} is not a standard online CL method as it retrains all replay-data with data augmentation for 256 epochs after each task, which is not suitable for online CL because it cannot be used for any-time inference. We thus removed this retraining operation and denote it as Rainbow-rt.} 





\subsection{Architecture, Data Augmentation, Training Details and Evaluation Protocol}
\label{sec.arch}
\textbf{Architecture}.
For MNIST, GSA and baselines employ a fully-connected network with two hidden layers as the feature extractor $h_{\theta}$, each comprising of 400 ReLU units. A linear layer of size [400, 10] is used as the classifier $f_\phi$.
For CIFAR10, CIFAR100, and TinyImageNet, we follow~\cite{buzzega2020dark} and use ResNet18 (not pre-trained) as the feature extractor $h_{\theta}$ {with around 11 million trainable parameters} for GSA and all baselines. Denoting $C_{\textit{num}}$ as the number of all classes, we employ a linear layer of size [$dim_h$, $C_{\textit{num}}$] as the classifier $f_\phi$. 
For an input $x$, we use $F(x)$ to compute $\mathcal{L}_{ce}$.

\textbf{Data Augmentation.} To learn better features, we apply two data augmentations to each image in $X^{\textit{new}}$ and $X^{\textit{mix}}$:  \textit{random-resized-crop} and \textit{random-gray-scale}. {\color{black}For fair comparison, the same data augmentations are applied to all baselines ($X^{\textit{new}}$ and $X^{\textit{buf}}$) to improve their performance, which results in an average of about 2\% of improvement for all methods with no drop in performance for any baseline.} 

\textbf{Training and hyperparameter settings}.
{Like ER and many other online CL systems, GSA uses reservoir sampling for memory update.} We follow~\cite{guo2022online} and train GSA with the Adam optimizer. We set the learning rate as 0.001 and fix the weight decay as 0.0001 for all settings. We follow~\cite{guo2022online} and set the batch size of $X^{\textit{new}}$ as 10 for all methods and $X^{\textit{mix}}$ as 64 for our GSA and $X^{\textit{buf}}$ also as 64 for baselines. 
We list other hyper-parameters in {\color{black}Appendix 6}.

We use the official codes of baselines. Their source links and default hyper-parameters are listed in {Appendix 6}. \textit{We run all methods with one epoch for each task}. 



\textbf{Evaluation protocol.} We first learn all tasks from the data stream for each dataset, and then test the final model using the test data of all tasks. We report the average accuracy of 15 random runs. 

\subsection{Results Analysis and Ablation Experiments}
\label{sec.results}
The results in  Table~\ref{tab:accresults} show that the best online CL baseline is SCR and the best class-imbalance baseline is BiC. Although BiC and SSIL were not originally designed for online CL, they perform well for online CL, better than all online CL baselines, which do not deal with data imbalance. This indicates that the data-balanced approaches in BiC and SSIL help. Our method GSA consistently outperforms them by very large margins as we also deal with the proposed CL imbalance. 
{\color{black}Note that OCM is not compared in the tables as it is not a competitor of GSA. GSA is actually complementary to OCM. 
For example, with the largest memory size for each dataset, 
OCM+GSA gives 96.5\% on MNIST, 77.5\% on CIFAR10, 53.7\% on CIFAR100, and 35.7\% on TinyImageNet and also outperforms OCM~\cite{guo2022online}.}

\textbf{Forgetting rate.} 
Table~\ref{tab:forgettingrate} shows that our GSA has substantially lower forgetting rates than baselines except for GDumb, Rainbow-rt, and SCR (in two datasets), but GDumb, Rainbow-rt, and SCR's accuracy values are substantially lower than that of GSA (see Table~\ref{tab:accresults}).
The calculation method of the average forgetting rate and \textbf{Training times} are given in Appendix 7.

{\color{black}\textbf{Ablation.}
We conducted ablation experiments on GSA using two datasets, CIFAR100 and TinyImageNet. The results are given in  Table~\ref{tab:ablation}. 

\textbf{(1).} \textit{Ablation study of training loss in GSA.}  
In the experiments ``no new loss'', "no previous loss", and "no cross loss", we do not consider the first term (establishing the boundaries between new classes), the second term (maintaining the decision boundaries between previous classes), and the last term (establishing the cross-task boundaries) in Eq.~\ref{eq:final} respectively. 
Table~\ref{tab:ablation} shows that their performances are all poorer than GSA (Table~\ref{tab:accresults}). This is because in the first case, the model forgets the knowledge of previous classes, which causes a drastic drop in  performance. 
{The second experiment shows that our method benefits from considering all the new data in $X^{\textit{new}}$.} 
The third experiment shows that establishing cross-task boundaries is an important problem for CIL and can improve the overall performance further.

\textbf{(2).} \textit{Ablation study of data imbalance and CL imbalance}.
In the experiments ``no balanced sampling,'' we replace the sampling strategy for $X^{\textit{mix}}$ with random sampling from the memory buffer $X^{\textit{buf}}$. In the experiment ``no CL imbalance,'' {\color{black}we do not use the $\mathcal{L}_{\textit{GSA-CE}}$ losses proposed in Eq.~\ref{eq:mix} and Eq.~\ref{eq:buf} and introduced into the final optimization objective in Eq.~\ref{eq:final}, 
but replace them with the simple cross-entropy loss. 
Table~\ref{tab:ablation} shows that all these incomplete GSA systems are poorer than the full GSA in Table~\ref{tab:accresults}. {\color{black}The poorer performance of ``no balanced-sampling'' is mainly due to data imbalance between new classes and previous classes in the buffer. Specifically, in the initial training process of the new task, the stored training data of the new task classes are usually fewer than that of the previous classes in the buffer and the cross-task class boundaries are not established well.} The poorer performance of ``no CL imbalance'' is because the method does not consider different gradient rates across different classes, which again results in poorer performances.}

\textbf{GSA-CE loss mitigates the gradient imbalance.} Fig.~\ref{Fig.7} shows that the GSA-CE loss clearly reduces the A-PN gradient rate imbalance than the method without it by dropping the A-NP rate that is bigger than -1 and improving the A-NP rate that is smaller than -1. And it makes the curve of A-PN rate more similar to the joint training curve.

\begin{figure}
\centering
\begin{minipage}[b]{1\textwidth} 
\scalebox{0.85}{\includegraphics[width=0.5\textwidth,height=0.325\textwidth]{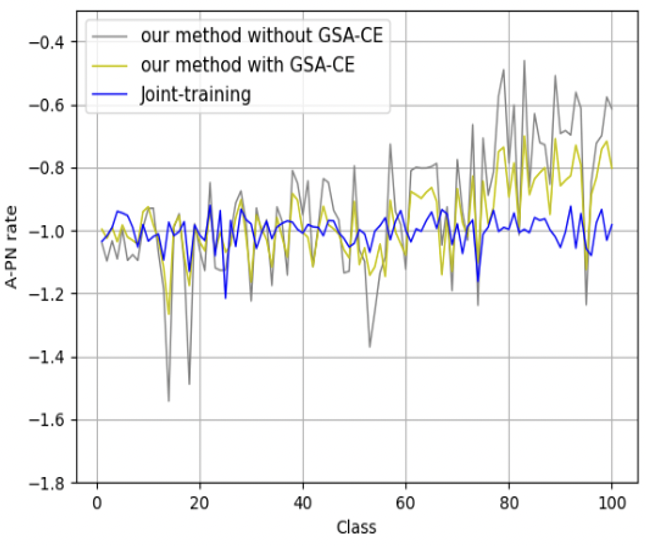}} 
\end{minipage}
\caption{The gradient rate after the model has learned all classes of CIFAR100. A-PN is the gradient rate. The memory buffer size is 1000. Joint training means the model learns all classes together as one task in one epoch. }
\label{Fig.7}
\end{figure}
}
{\color{black}\textbf{Long-tail online CL experiments.} See Appendix 8, where we will show that our
method also outperforms the baselines. 
\section{Conclusion}
{\color{black}This paper discussed the challenge of \textit{cross-task class discrimination} (CTCD) and showed that the replay approach partially deals with the problem. However, the replay approach has a major bias in training, which is manifested by \textit{gradient imbalance on the logits} and significantly limits the online CL performance and the ability to solve the CTCD problem. The paper then analyzed gradient imbalance from two perspectives: \textit{data imbalance} introduced by replay and \textit{CL imbalance} due to CL itself. After that, it proposed a new learning strategy and a new self-adaptive loss function to deal with the problems. Empirical evaluation demonstrated that the new approach GSA improves the baselines by large margins. 
\thispagestyle{empty}
\appendix

{\small
\bibliographystyle{ieee_fullname}
\bibliography{egbib}

\begin{thebibliography}{10}\itemsep=-1pt

\bibitem{abati2020conditional}
Davide Abati, Jakub Tomczak, Tijmen Blankevoort, Simone Calderara, Rita
  Cucchiara, and Babak~Ehteshami Bejnordi.
\newblock Conditional channel gated networks for task-aware continual learning.
\newblock In {\em Proceedings of the IEEE/CVF Conference on Computer Vision and
  Pattern Recognition}, pages 3931--3940, 2020.

\bibitem{ahn2021ss}
Hongjoon Ahn, Jihwan Kwak, Subin Lim, Hyeonsu Bang, Hyojun Kim, and Taesup
  Moon.
\newblock Ss-il: Separated softmax for incremental learning.
\newblock In {\em Proceedings of the IEEE/CVF International Conference on
  Computer Vision}, pages 844--853, 2021.

\bibitem{aljundi2019online}
Rahaf Aljundi, Lucas Caccia, Eugene Belilovsky, Massimo Caccia, Min Lin,
  Laurent Charlin, and Tinne Tuytelaars.
\newblock Online continual learning with maximally interfered retrieval.
\newblock {\em arXiv preprint arXiv:1908.04742}, 2019.

\bibitem{aljundi2019gradient}
Rahaf Aljundi, Min Lin, Baptiste Goujaud, and Yoshua Bengio.
\newblock Gradient based sample selection for online continual learning.
\newblock {\em arXiv preprint arXiv:1903.08671}, 2019.

\bibitem{bang2021rainbow}
Jihwan Bang, Heesu Kim, YoungJoon Yoo, Jung-Woo Ha, and Jonghyun Choi.
\newblock Rainbow memory: Continual learning with a memory of diverse samples.
\newblock In {\em Proceedings of the IEEE/CVF Conference on Computer Vision and
  Pattern Recognition}, pages 8218--8227, 2021.

\bibitem{buzzega2020dark}
Pietro Buzzega, Matteo Boschini, Angelo Porrello, Davide Abati, and Simone
  Calderara.
\newblock Dark experience for general continual learning: a strong, simple
  baseline.
\newblock {\em arXiv preprint arXiv:2004.07211}, 2020.

\bibitem{caccia2021reducing}
Lucas Caccia, Rahaf Aljundi, Nader Asadi, Tinne Tuytelaars, Joelle Pineau, and
  Eugene Belilovsky.
\newblock Reducing representation drift in online continual learning.
\newblock {\em arXiv preprint arXiv:2104.05025}, 2021.

\bibitem{castro2018end}
Francisco~M Castro, Manuel~J Mar{\'\i}n-Jim{\'e}nez, Nicol{\'a}s Guil, Cordelia
  Schmid, and Karteek Alahari.
\newblock End-to-end incremental learning.
\newblock In {\em ECCV}, 2018.

\bibitem{Cha_2021_ICCV}
Hyuntak Cha, Jaeho Lee, and Jinwoo Shin.
\newblock Co2l: Contrastive continual learning.
\newblock In {\em Proceedings of the IEEE/CVF International Conference on
  Computer Vision (ICCV)}, pages 9516--9525, October 2021.

\bibitem{chaudhry2018efficient}
Arslan Chaudhry, Marc'Aurelio Ranzato, Marcus Rohrbach, and Mohamed Elhoseiny.
\newblock Efficient lifelong learning with a-gem.
\newblock {\em arXiv preprint arXiv:1812.00420}, 2018.

\bibitem{chaudhry2019continual}
Arslan Chaudhry, Marcus Rohrbach, Mohamed Elhoseiny, Thalaiyasingam Ajanthan,
  Puneet~K Dokania, Philip~HS Torr, and M Ranzato.
\newblock Continual learning with tiny episodic memories.
\newblock In {\em ICML-2019}, 2019.

\bibitem{chaudhry2019tiny}
Arslan Chaudhry, Marcus Rohrbach, Mohamed Elhoseiny, Thalaiyasingam Ajanthan,
  Puneet~K Dokania, Philip~HS Torr, and Marc'Aurelio Ranzato.
\newblock On tiny episodic memories in continual learning.
\newblock {\em arXiv preprint arXiv:1902.10486}, 2019.

\bibitem{Cyprien2019episodic}
Cyprien de Masson~d’Autume, Sebastian Ruder, Lingpeng Kong, and Dani
  Yogatama.
\newblock Episodic memory in lifelong language learning.
\newblock In {\em NeurIPS}, 2019.

\bibitem{farajtabar2020orthogonal}
Mehrdad Farajtabar, Navid Azizan, Alex Mott, and Ang Li.
\newblock Orthogonal gradient descent for continual learning.
\newblock In {\em International Conference on Artificial Intelligence and
  Statistics}, pages 3762--3773. PMLR, 2020.

\bibitem{guo2022adaptive}
Yiduo Guo, Wenpeng Hu, Dongyan Zhao, and Bing Liu.
\newblock Adaptive orthogonal projection for batch and online continual
  learning.
\newblock In {\em Proceedings of AAAI-2021}, 2022.

\bibitem{guo2022online}
Yiduo Guo, Bing Liu, and Dongyan Zhao.
\newblock Online continual learning through mutual information maximization.
\newblock In {\em International Conference on Machine Learning}, pages
  8109--8126. PMLR, 2022.

\bibitem{guo2020improved}
Yunhui Guo, Mingrui Liu, Tianbao Yang, and Tajana Rosing.
\newblock Improved schemes for episodic memory-based lifelong learning.
\newblock {\em Advances in Neural Information Processing Systems},
  33:1023--1035, 2020.

\bibitem{hayes2019remind}
Tyler~L Hayes, Kushal Kafle, Robik Shrestha, Manoj Acharya, and Christopher
  Kanan.
\newblock Remind your neural network to prevent catastrophic forgetting.
\newblock {\em arXiv preprint arXiv:1910.02509}, 2019.

\bibitem{hou2019learning}
Saihui Hou, Xinyu Pan, Chen~Change Loy, Zilei Wang, and Dahua Lin.
\newblock Learning a unified classifier incrementally via rebalancing.
\newblock In {\em CVPR}, pages 831--839, 2019.

\bibitem{hung2019neurIPS}
Steven C.~Y. Hung, Cheng-Hao Tu, Cheng-En Wu, Chien-Hung Chen, Yi-Ming Chan,
  and Chu-Song Chen.
\newblock Compacting, picking and growing for unforgetting continual learning.
\newblock In {\em NeurIPS}, 2019.

\bibitem{Kemker2018fearnet}
Ronald Kemker and Christopher Kanan.
\newblock {FearNet: Brain-Inspired Model for Incremental Learning}.
\newblock In {\em ICLR}, 2018.

\bibitem{kimtheoretical}
Gyuhak Kim, Changnan Xiao, Tatsuya Konishi, Zixuan Ke, and Bing Liu.
\newblock A theoretical study on solving continual learning.
\newblock In {\em Advances in Neural Information Processing Systems}, 2022.

\bibitem{kirkpatrick2017overcoming}
James Kirkpatrick, Razvan Pascanu, Neil Rabinowitz, Joel Veness, Guillaume
  Desjardins, Andrei~A Rusu, Kieran Milan, John Quan, Tiago Ramalho, Agnieszka
  Grabska-Barwinska, et~al.
\newblock Overcoming catastrophic forgetting in neural networks.
\newblock {\em Proceedings of the national academy of sciences},
  114(13):3521--3526, 2017.

\bibitem{korycki2021class}
{\L}ukasz Korycki and Bartosz Krawczyk.
\newblock Class-incremental experience replay for continual learning under
  concept drift.
\newblock {\em arXiv preprint arXiv:2104.11861}, 2021.

\bibitem{Krizhevsky2009learning}
Alex Krizhevsky and Geoffrey Hinton.
\newblock Learning multiple layers of features from tiny images.
\newblock {\em Technical Report TR-2009, University of Toronto, Toronto.},
  2009.

\bibitem{le2015tiny}
Ya Le and Xuan Yang.
\newblock Tiny imagenet visual recognition challenge.
\newblock {\em CS 231N}, 7:7, 2015.

\bibitem{leCun1998mnist}
Yann LeCun, Corinna Cortes, and Christopher~JC Burges.
\newblock The mnist database of handwritten digits.
\newblock {\em http://yann.lecun.com/exdb/mnist/}, 1998.

\bibitem{li2019learn}
Xilai Li, Yingbo Zhou, Tianfu Wu, Richard Socher, and Caiming Xiong.
\newblock Learn to grow: A continual structure learning framework for
  overcoming catastrophic forgetting.
\newblock In {\em ICML}, 2019.

\bibitem{Lopez2017gradient}
David Lopez-Paz and Marc’Aurelio Ranzato.
\newblock {Gradient Episodic Memory for Continual Learning}.
\newblock In {\em NIPS}, pages 6470--6479, 2017.

\bibitem{mai2021supervised}
Zheda Mai, Ruiwen Li, Hyunwoo Kim, and Scott Sanner.
\newblock Supervised contrastive replay: Revisiting the nearest class mean
  classifier in online class-incremental continual learning.
\newblock In {\em Proceedings of the IEEE/CVF Conference on Computer Vision and
  Pattern Recognition}, pages 3589--3599, 2021.

\bibitem{Mai_2021_CVPR}
Zheda Mai, Ruiwen Li, Hyunwoo Kim, and Scott Sanner.
\newblock Supervised contrastive replay: Revisiting the nearest class mean
  classifier in online class-incremental continual learning.
\newblock In {\em Proceedings of the IEEE/CVF Conference on Computer Vision and
  Pattern Recognition (CVPR) Workshops}, pages 3589--3599, 2021.

\bibitem{masana2020class}
Marc Masana, Xialei Liu, Bartlomiej Twardowski, Mikel Menta, Andrew~D Bagdanov,
  and Joost van~de Weijer.
\newblock Class-incremental learning: survey and performance evaluation on
  image classification.
\newblock {\em arXiv preprint arXiv:2010.15277}, 2020.

\bibitem{mittal2021essentials}
Sudhanshu Mittal, Silvio Galesso, and Thomas Brox.
\newblock Essentials for class incremental learning.
\newblock In {\em Proceedings of the IEEE/CVF Conference on Computer Vision and
  Pattern Recognition}, pages 3513--3522, 2021.

\bibitem{ostapenko2019learning}
Oleksiy Ostapenko, Mihai Puscas, Tassilo Klein, Patrick Jahnichen, and Moin
  Nabi.
\newblock Learning to remember: A synaptic plasticity driven framework for
  continual learning.
\newblock In {\em CVPR}, pages 11321--11329, 2019.

\bibitem{prabhu2020gdumb}
Ameya Prabhu, Philip~HS Torr, and Puneet~K Dokania.
\newblock Gdumb: A simple approach that questions our progress in continual
  learning.
\newblock In {\em EECV}, pages 524--540, 2020.

\bibitem{rajasegaran2019neurIPS}
Jathushan Rajasegaran, Munawar Hayat, Salman Khan, Fahad Shahbaz, and Khan~Ling
  Shao.
\newblock Random path selection for incremental learning.
\newblock In {\em NeurIPS}, 2019.

\bibitem{rebuffi2017icarl}
Sylvestre-Alvise Rebuffi, Alexander Kolesnikov, Georg Sperl, and Christoph~H
  Lampert.
\newblock icarl: Incremental classifier and representation learning.
\newblock In {\em CVPR}, pages 2001--2010, 2017.

\bibitem{Seff2017continual}
Ari Seff, Alex Beatson, Daniel Suo, and Han Liu.
\newblock {Continual learning in generative adversarial nets}.
\newblock {\em arXiv preprint arXiv:1705.08395}, 2017.

\bibitem{Serra2018overcoming}
Joan Serr{\`{a}}, D{\'{i}}dac Sur{\'{i}}s, Marius Miron, and Alexandros
  Karatzoglou.
\newblock {Overcoming catastrophic forgetting with hard attention to the task}.
\newblock In {\em ICML}, 2018.

\bibitem{shim2021online}
Dongsub Shim, Zheda Mai, Jihwan Jeong, Scott Sanner, Hyunwoo Kim, and Jongseong
  Jang.
\newblock Online class-incremental continual learning with adversarial shapley
  value.
\newblock In {\em Proceedings of the AAAI Conference on Artificial Intelligence
  (AAAI)}, pages 9630--9638, 2021.

\bibitem{Shin2017continual}
Hanul Shin, Jung~Kwon Lee, Jaehong Kim, and Jiwon Kim.
\newblock Continual learning with deep generative replay.
\newblock In {\em NIPS}, pages 2994--3003, 2017.

\bibitem{soutif2021importance}
Albin Soutif-Cormerais, Marc Masana, Joost Van~de Weijer, and Bartl{\o}miej
  Twardowski.
\newblock On the importance of cross-task features for class-incremental
  learning.
\newblock {\em arXiv: 2106.11930}, 2021.

\bibitem{von2019continual}
Johannes von Oswald, Christian Henning, Jo{\~a}o Sacramento, and Benjamin~F
  Grewe.
\newblock Continual learning with hypernetworks.
\newblock {\em ICLR}, 2020.

\bibitem{wang2022memory}
Liyuan Wang, Xingxing Zhang, Kuo Yang, Longhui Yu, Chongxuan Li, Lanqing Hong,
  Shifeng Zhang, Zhenguo Li, Yi Zhong, and Jun Zhu.
\newblock Memory replay with data compression for continual learning.
\newblock {\em ICLR-2022}, 2022.

\bibitem{wu2018memory}
Chenshen Wu, Luis Herranz, Xialei Liu, Joost van~de Weijer, Bogdan Raducanu,
  et~al.
\newblock Memory replay gans: Learning to generate new categories without
  forgetting.
\newblock In {\em NIPS}, pages 5962--5972, 2018.

\bibitem{wu2019large}
Yue Wu, Yinpeng Chen, Lijuan Wang, Yuancheng Ye, Zicheng Liu, Yandong Guo, and
  Yun Fu.
\newblock Large scale incremental learning.
\newblock In {\em Proceedings ofthe IEEE/CVF Conference on Computer Vision and
  Pattern Recognition (CVPR)}, 2019.

\bibitem{yan2021framework}
Shipeng Yan, Jiale Zhou, Jiangwei Xie, Songyang Zhang, and Xuming He.
\newblock An em framework for online incremental learning of semantic
  segmentation.
\newblock In {\em Proceedings of the 29th ACM International Conference on
  Multimedia}, pages 3052--3060, 2021.

\bibitem{yin2021mitigating}
Haiyan Yin, Ping Li, et~al.
\newblock Mitigating forgetting in online continual learning with neuron
  calibration.
\newblock {\em Advances in Neural Information Processing Systems}, 34, 2021.

\bibitem{Yoon2018lifelong}
Jaehong Yoon, Eunho Yang, Jeongtae Lee, and Sung~Ju Hwang.
\newblock {Lifelong Learning with Dynamically Expandable Networks}.
\newblock In {\em ICLR}, 2018.

\bibitem{yu2020semantic}
Lu Yu, Bartlomiej Twardowski, Xialei Liu, Luis Herranz, Kai Wang, Yongmei
  Cheng, Shangling Jui, and Joost van~de Weijer.
\newblock Semantic drift compensation for class-incremental learning.
\newblock In {\em Proceedings of the IEEE/CVF Conference on Computer Vision and
  Pattern Recognition}, pages 6982--6991, 2020.

\bibitem{zeng2019continual}
Guanxiong Zeng, Yang Chen, Bo Cui, and Shan Yu.
\newblock Continual learning of context-dependent processing in neural
  networks.
\newblock {\em Nature Machine Intelligence}, 1(8):364--372, 2019.

\bibitem{zhang2020class}
Junting Zhang, Jie Zhang, Shalin Ghosh, Dawei Li, Serafettin Tasci, Larry Heck,
  Heming Zhang, and C.-C.~Jay Kuo.
\newblock Class-incremental learning via deep model consolidation.
\newblock In {\em CVPR}, 2020.

\bibitem{zhao2020maintaining}
Bowen Zhao, Xi Xiao, Guojun Gan, Bin Zhang, and Shu-Tao Xia.
\newblock Maintaining discrimination and fairness in class incremental
  learning.
\newblock In {\em Proceedings of the IEEE/CVF Conference on Computer Vision and
  Pattern Recognition}, pages 13208--13217, 2020.

\bibitem{zhao2021memory}
Hanbin Zhao, Hui Wang, Yongjian Fu, Fei Wu, and Xi Li.
\newblock Memory efficient class-incremental learning for image classification.
\newblock {\em IEEE Transactions on Neural Networks and Learning Systems},
  2021.

\bibitem{zhu2021class}
Fei Zhu, Zhen Cheng, Xu-yao Zhang, and Cheng-lin Liu.
\newblock Class-incremental learning via dual augmentation.
\newblock {\em Advances in Neural Information Processing Systems}, 34, 2021.

\end{thebibliography}
}

\end{document}